\documentclass[11pt,a4paper]{article}
\usepackage{acl2015}
\usepackage{times}
\usepackage{url}
\usepackage{latexsym}
\usepackage{amsmath}
\usepackage{amsfonts}

\title{Better Summarization Evaluation with Word Embeddings for ROUGE}

\author{Jun-Ping Ng \\
  Bloomberg L.P. \\ 
  New York, USA \\
  {\tt jng324@bloomberg.net}  \\\And
  Viktoria Abrecht \\
  Bloomberg L.P. \\
  New York, USA \\
  {\tt vkanchakousk@bloomberg.net} 
\\}

\date{}

\begin{document}
\maketitle
\begin{abstract}
ROUGE is a widely adopted, automatic evaluation measure for text summarization.
While it has been shown to correlate well with human judgements, it is biased towards surface lexical similarities. 
This makes it unsuitable for the evaluation of abstractive summarization, or summaries with substantial paraphrasing. 
We study the effectiveness of word embeddings to overcome this disadvantage of ROUGE.   Specifically, instead of measuring lexical overlaps, word embeddings are used to compute the semantic similarity of the words used in summaries instead. 
Our experimental results show that our proposal is able to achieve better correlations with human judgements when measured with the Spearman and Kendall rank coefficients.
\end{abstract}

\section{Introduction}
\label{sec:introduction}
Automatic text summarization is a rich field of research.
For example, shared task evaluation workshops for summarization were held for more than a decade in the Document Understanding Conference (DUC), and subsequently the Text Analysis Conference (TAC).
An important element of these shared tasks is the evaluation of participating systems.
Initially, manual evaluation was carried out, where human judges were tasked to assess the quality of automatically generated summaries.
However in an effort to make evaluation more scaleable, the automatic ROUGE\footnote{Recall-Oriented Understudy of Gisting Evaluation} measure~\cite{lin2004_rouge} was introduced in DUC-2004.
ROUGE determines the quality of an automatic summary through comparing overlapping units such as n-grams, word sequences, and word pairs with human written summaries. 

ROUGE is not perfect however. 
Two problems with ROUGE are that 1) it favors lexical similarities between generated summaries and model summaries, which makes it unsuitable to evaluate abstractive summarization, or summaries with a significant amount of paraphrasing, and 2) it does not make any provision to cater for the readability or fluency of the generated summaries.

There has been on-going efforts to improve on automatic summarization evaluation measures, such as the Automatically Evaluating Summaries of Peers (AESOP) task in TAC~\cite{dang2009summ,owczarzak2010summ,owczarzak2011summ}.
However, ROUGE remains as one of the most popular metric of choice, as it has repeatedly been shown to correlate very well with human judgements~\cite{lin2004rougeevaluation,over2004duc,owczarzak2011summ}.

In this work, we describe our efforts to tackle the first problem of ROUGE that we have identified above --- its bias towards lexical similarities. 
We propose to do this by making use of word embeddings~\cite{bengio2003_we}. 
Word embeddings refer to the mapping of words into a multi-dimensional vector space.
We can construct the mapping, such that the distance between two word projections in the vector space corresponds to the semantic similarity between the two words.
By incorporating these word embeddings into ROUGE, we can overcome its bias towards lexical similarities and instead make comparisons based on the semantics of words sequences.
We believe that this will result in better correlations with human assessments, and avoid situations where two word sequences share similar meanings, but get unfairly penalized by ROUGE due to differences in lexicographic representations.

As an example, consider these two phrases: 1) \emph{It is raining heavily}, and 2) \emph{It is pouring}.
If we are performing a lexical string match, as ROUGE does, there is nothing in common between the terms ``raining'', ``heavily'', and ``pouring''. 
However, these two phrases mean the same thing.
If one of the phrases was part of a human written summary, while the other was output by an automatic summarization system, we want to be able to reward the automatic system accordingly.

%
%
%

In our experiments, we show that word embeddings indeed give us better correlations with human judgements when measured with the Spearman and Kendall rank coefficient.
This is a significant and exciting result. 
Beyond just improving the evaluation prowess of ROUGE, it has the potential to expand the applicability of ROUGE to abstractive summmarization as well.

\section{Related Work}
\label{sec:related_work}
While ROUGE is widely-used, as we have noted earlier, there is a significant body of work studying the evaluation of automatic text summarization systems.
A good survey of many of these measures has been written by \newcite{steinberger2012evaluation}.
We will thus not attempt to go through every measure here, but rather highlight the more significant efforts in this area.

Besides ROUGE, Basic Elements (BE)~\cite{hovy2005basicelements} has also been used in the DUC/TAC shared task evaluations. 
It is an automatic method which evaluates the content completeness of a generated summary by breaking up sentences into smaller, more granular units of information (referred to as ``Basic Elements'').

The pyramid method originally proposed by \newcite{passonneau2005_pyramid} is another staple in DUC/TAC. 
However it is a semi-automated method, where significant human intervention is required to identify units of information, called \emph{Summary Content Units} (SCUs), and then to map content within generated summaries to these SCUs.
Recently however, an automated variant of this method has been proposed~\cite{passonneau2013automated}.
In this variant, word embeddings are used, as we are proposing in this paper, to map text content within generated summaries to SCUs.
However the SCUs still need to be manually identified, limiting this variant's scalability and applicability.

Many systems have also been proposed in the AESOP task in TAC from 2009 to 2011.
For example, the top system reported in \newcite{owczarzak2011summ}, AutoSummENG~\cite{giannakopoulos2009_autosummeng}, is a graph-based system which scores summaries based on the similarity between the graph structures of the generated summaries and model summaries.

\section{Methodology}
\label{sec:methodology}
Let us now describe our proposal to integrate word embeddings into ROUGE in greater detail.

To start off, we will first describe the word embeddings that we intend to adopt.
A word embedding is really a function $W$, where $W : w \rightarrow \mathbb{R}^{n} $, and $w$ is a word or word sequence.
For our purpose, we want $W$ to map two words $w_1$ and $w_2$ such that their respective projections are closer to each other if the words are semantically similar, and further apart if they are not.
\newcite{mikolov2013linguistic} describe one such variant, called \texttt{word2vec}, which gives us this desired property\footnote{The effectiveness of the learnt mapping is such that we can now compute analogies such as \emph{king} $-$ \emph{man} $+$ \emph{woman} $=$ \emph{queen}.}. 
We will thus be making use of \texttt{word2vec}.

We will now explain how word embeddings can be incorporated into ROUGE.
There are several variants of ROUGE, of which ROUGE-1, ROUGE-2, and ROUGE-SU4 have often been used.
This is because they have been found to correlate well with human judgements~\cite{lin2004rougeevaluation,over2004duc,owczarzak2011summ}.
ROUGE-1 measures the amount of unigram overlap between model summaries and automatic summaries, and ROUGE-2 measures the amount of bigram overlap. 
ROUGE-SU4 measures the amount of overlap of skip-bigrams, which are pairs of words in the same order as they appear in a sentence.
In each of these variants, overlap is computed by matching the lexical form of the words within the target pieces of text.
Formally, we can define this as a similarity function $f_{R}$ such that:
\begin{equation}
f_{R}( w_1, w_2 ) = 
\begin{cases}
	1,& \text{if } w_1 = w_2 \\
	0,& \text{otherwise}
\end{cases}
\end{equation}
where $w_1$ and $w_2$ are the words (could be unigrams or n-grams) being compared.

In our proposal\footnote{\url{https://github.com/ng-j-p/rouge-we}}, which we will refer to as ROUGE-WE, we define a new similarity function $f_{WE}$ such that:
\begin{equation}
f_{WE}( w_1, w_2 ) = 
\begin{cases}
	0, \text{ if } v_1 \text{or } v_2 \text{ are OOV}\\
	v_1 \cdot v_2, \text{ otherwise} \\
\end{cases}
\end{equation}
where $w_1$ and $w_2$ are the words being compared, and $v_x = W(w_x)$. 
\emph{OOV} here means a situation where we encounter a word $w$ that our word embedding function $W$ returns no vector for.
For the purpose of this work, we make use of a set of 3 million pre-trained vector mappings\footnote{\url{https://drive.google.com/file/d/0B7XkCwpI5KDYNlNUTTlSS21pQmM/edit?usp=sharing}} trained from part of Google's news dataset~\cite{mikolov2013_distributed} for $W$.

\noindent\textbf{Reducing OOV terms for n-grams.} 
With our formulation for $f_{WE}$, we are able to compute variants of ROUGE-WE that correspond to those of ROUGE, including ROUGE-WE-1, ROUGE-WE-2, and ROUGE-WE-SU4.
However, despite the large number of vector mappings that we have, there will still be a large number of OOV terms in the case of ROUGE-WE-2 and ROUGE-WE-SU4, where the basic units of comparison are bigrams.

To solve this problem, we can compose individual word embeddings together.
We follow the simple multiplicative approach described by \newcite{mitchell2008_compose}, where individual vectors of constituent tokens are multiplied together to produce the vector for a n-gram, \emph{i.e.},
\begin{equation}
W(w) = W(w_1) \times \ldots \times W(w_n)
\end{equation}
where $w$ is a n-gram composed of individual word tokens, \emph{i.e.}, $w = w_1 w_2 \ldots w_n$.
Multiplication between two vectors $W(w_i) = \{v_{i1}, \ldots, v_{ik}\}$ and $W(w_j) = \{v_{j1}, \ldots, v_{jk}\}$ in this case is defined as:
\begin{equation}
\{v_{i1} \times v_{j1}, \ldots, v_{ik} \times v_{jk}\}
\end{equation}


\section{Experiments}
\label{sec:experiments}
\subsection{Dataset and Metrics}

For our experiments, we make use of the dataset used in AESOP~\cite{owczarzak2011summ}, and the corresponding correlation measures.

For clarity, let us first describe the dataset used in the main TAC summarization task.
The main summarization dataset consists of 44 topics, each of which is associated with a set of 10 documents.
There are also four human-curated model summaries for each of these topics.
Each of the 51 participating systems generated a summary for each of these topics.
These automatically generated summaries, together with the human-curated model summaries, then form the basis of the dataset for AESOP.

To assess how effective an automatic evaluation system is, the system is first tasked to assign a score for each of the summaries generated by all of the 51 participating systems.
Each of these summaries would also have been assessed by human judges using these three key metrics:

\noindent\textbf{Pyramid.} As reviewed in Section~\ref{sec:related_work}, this is a semi-automated measure described in \newcite{passonneau2005_pyramid}.

\noindent\textbf{Responsiveness.} Human judges are tasked to evaluate how well a summary adheres to the information requested, as well as the linguistic quality of the generated summary.

\noindent\textbf{Readability.} Human judges give their judgement on how fluent and readable a summary is.


The evaluation system's scores are then tested to see how well they correlate with the human assessments.
The correlation is evaluated with a set of three metrics, including 1) Pearson correlation (P), 2) Spearman rank coefficient (S), and 3) Kendall rank coefficient (K).


\subsection{Results}

We evaluate three different variants of our proposal, ROUGE-WE-1, ROUGE-WE-2, and ROUGE-WE-SU4, against their corresponding variants of ROUGE (\emph{i.e.}, ROUGE-1, ROUGE-2, ROUGE-SU4).
It is worth noting here that in AESOP in 2011, ROUGE-SU4 was shown to correlate very well with human judgements, especially for pyramid and responsiveness, and out-performs most of the participating systems.

Tables~\ref{tab:results_main_pyr}, \ref{tab:results_main_res}, and \ref{tab:results_main_fl} show the correlation of the scores produced by each variant of ROUGE-WE with human assessed scores for pyramid, responsiveness, and readability respectively. 
The tables also show the correlations achieved by ROUGE-1, ROUGE-2, and ROUGE-SU4.
The best result for each column has been bolded for readability.

\begin{table}[h]
\begin{center}
\begin{tabular}{|c|c|c|c|}
\hline \bf Measure & \bf P & \bf S & \bf K \\ \hline\hline
ROUGE-WE-1        &     0.9492  & \bf 0.9138   & \bf 0.7534  \\
ROUGE-WE-2        &     0.9765  &     0.8984   &     0.7439  \\
ROUGE-WE-SU4      &     0.9783  &     0.8808   &     0.7198  \\
ROUGE-1           &     0.9661  &     0.9085   &     0.7466  \\ 
ROUGE-2           &     0.9606  &     0.8943   &     0.7450  \\ 
ROUGE-SU4         & \bf 0.9806  &     0.8935   &     0.7371  \\ 
\hline
\end{tabular}
\end{center}
\caption{\label{tab:results_main_pyr} Correlation with pyramid scores, measured with Pearson $r$ (P), Spearman $\rho$ (S), and Kendall $\tau$ (K) coefficients.}
\end{table}

\begin{table}[h]
\begin{center}
\begin{tabular}{|c|c|c|c|}
\hline \bf Measure & \bf P & \bf S & \bf K \\ \hline\hline
ROUGE-WE-1        &    0.9155   & \bf  0.8192  &     0.6308  \\
ROUGE-WE-2        &    0.9534   &      0.7974  &     0.6149  \\
ROUGE-WE-SU4      &    0.9538   &      0.7872  &     0.5969  \\
ROUGE-1           &    0.9349   &      0.8182  & \bf 0.6334  \\ 
ROUGE-2           &    0.9416   &      0.7897  &     0.6096  \\ 
ROUGE-SU4         &\bf 0.9545   &      0.7902  &     0.6017  \\ 
\hline
\end{tabular}
\end{center}
\caption{\label{tab:results_main_res} Correlation with responsiveness scores, measured with Pearson $r$ (P), Spearman $\rho$ (S), and Kendall $\tau$ (K) coefficients.}
\end{table}

\begin{table}[h]
\begin{center}
\begin{tabular}{|c|c|c|c|}
\hline \bf Measure & \bf P & \bf S & \bf K \\ \hline\hline
ROUGE-WE-1        &     0.7846  & \bf 0.4312   & \bf 0.3216  \\
ROUGE-WE-2        &     0.7819  &     0.4141   &     0.3042  \\
ROUGE-WE-SU4      & \bf 0.7931  &     0.4068   &     0.3020  \\
ROUGE-1           &     0.7900  &     0.3914   &     0.2846  \\ 
ROUGE-2           &     0.7524  &     0.3975   &     0.2925  \\ 
ROUGE-SU4         &     0.7840  &     0.3953   &     0.2925  \\ 
\hline
\end{tabular}
\end{center}
\caption{\label{tab:results_main_fl} Correlation with readability scores, measured with Pearson $r$ (P), Spearman $\rho$ (S), and Kendall $\tau$ (K) coefficients.}
\end{table}


ROUGE-WE-1 is observed to correlate very well with the pyramid, responsiveness, and readability scores when measured with the Spearman and Kendall rank correlation.
However, ROUGE-SU4 correlates better with human assessments for the Pearson correlation.
The key difference between the Pearson correlation and Spearman/Kendall rank correlation, is that the former assumes that the variables being tested are normally distributed.
It also further assumes that the variables are linearly related to each other.
The latter two measures are however non-parametric and make no assumptions about the distribution of the variables being tested. 
We argue that the assumptions made by the Pearson correlation may be too constraining, given that any two independent evaluation systems may not exhibit linearity.

Looking at the two bigram based variants, ROUGE-WE-2 and ROUGE-WE-SU4, we observe that ROUGE-WE-2 improves on ROUGE-2 most of the time, regardless of the correlation metric used.
This lends further support to our proposal to use word embeddings with ROUGE.

However ROUGE-WE-SU4 is only better than ROUGE-SU4 when evaluating readability. 
It does consistently worse than ROUGE-SU4 for pyramid and responsiveness.
The reason for this is likely due to how we have chosen to compose unigram word vectors into bigram equivalents.
The multiplicative approach that we have taken worked better for ROUGE-WE-2 which looks at contiguous bigrams.
These are easier to interpret semantically than skip-bigrams (the target of ROUGE-WE-SU4).
The latter, by nature of their construction, loses some of the semantic meaning attached to each word, and thus may not be as amenable to the linear composition of word vectors.

\newcite{owczarzak2011summ} reports only the results of the top systems in AESOP in terms of Pearson's correlation. 
To get a more complete picture of the usefulness of our proposal, it will be instructive to also compare it against the other top systems in AESOP, when measured with the Spearman/Kendall correlations.
We show in Table~\ref{tab:results_sup_py} the top three systems which correlate best with the pyramid score when measured with the Spearman rank coefficient.
\texttt{C\_S\_IIITH3}~\cite{kumar2011_csiiith} is a graph-based system which assess summaries based on differences in word co-locations between generated summaries and model summaries.
\texttt{BE-HM} (baseline by the organizers of the AESOP task) is the BE system~\cite{hovy2005basicelements}, where basic elements are identified using a head-modifier criterion on parse results from Minipar.
Lastly, \texttt{catolicasc1}~\cite{oliverira2011_catolicasc} is also a graph-based system which frames the summary evaluation problem as a maximum bipartite graph matching problem.

\begin{table}[h]
\begin{center}
\begin{tabular}{|c|c|c|c|}
\hline \bf Measure & \bf S       & \bf K  \\ \hline\hline
ROUGE-WE-1        & \bf 0.9138  &     0.7534 \\ 
C\_S\_IIITH3      &     0.9033  & \bf 0.7582 \\
BE-HM             &     0.9030  &     0.7456 \\
catolicasc1       &     0.9017  &     0.7351 \\
\hline
\end{tabular}
\end{center}
\caption{\label{tab:results_sup_py} Correlation with pyramid scores of top systems in AESOP 2011, measured with the Spearman $\rho$ (S), and Kendall $\tau$ (K) coefficients.}
\end{table}

We see that ROUGE-WE-1 displays better correlations with pyramid scores than the top system in AESOP 2011 (\emph{i.e.}, \texttt{C\_S\_IIITH3}) when measured with the Spearman coefficient. 
The latter does slightly better however for the Kendall coefficient.
This observation further validates that our proposal is an effective enhancement to ROUGE.


%
%


%
%


\section{Conclusion}
We proposed an enhancement to the popular ROUGE metric in this work, ROUGE-WE.
ROUGE is biased towards identifying lexical similarity when assessing the quality of a generated summary.
We improve on this by incorporating the use of word embeddings.
This enhancement allows us to go beyond surface lexicographic matches, and capture instead the semantic similarities between words used in a generated summary and a human-written model summary.
Experimenting on the TAC AESOP dataset, we show that this proposal exhibits very good correlations with human assessments, measured with the Spearman and Kendall rank coefficients.
In particular, ROUGE-WE-1 outperforms leading state-of-the-art systems consistently.

Looking ahead, we want to continue building on this work.
One area to improve on is the use of a more inclusive evaluation dataset.
The AESOP summaries that we have used in our experiments are drawn from systems participating in the TAC summarization task, where there is a strong exhibited bias towards extractive summarizers.
It will be helpful to enlarge this set of summaries to include output from summarizers which carry out substantial paraphrasing~\cite{li2013document,ng2014_timeline,liu2015_abstractive}.

Another immediate goal is to study the use of better compositional embedding models.
The generalization of unigram word embeddings into bigrams (or phrases), is still an open problem~\cite{yin2014_we,yu2014_compositionalwe}.
A better compositional embedding model than the one that we adopted in this work should help us improve the results achieved by bigram variants of ROUGE-WE, especially ROUGE-WE-SU4.
This is important because earlier works have demonstrated the value of using skip-bigrams for summarization evaluation.

An effective and accurate automatic evaluation measure will be a big boon to our quest for better text summarization systems. 
Word embeddings add a promising dimension to summarization evaluation, and we hope to expand on the work we have shared  to further realize its potential.


\bibliographystyle{acl}
\bibliography{900_references}

\end{document}